# Measuring the Meaning of Words in Contexts: An automated analysis of controversies about 'Monarch butterflies,' 'Frankenfoods,' and 'stem cells'

Loet Leydesdorff [i] and Iina Hellsten [ii]

**Abstract**

Co-words have been considered as carriers of meaning across different domains in studies of science, technology, and society. Words and co-words, however, obtain meaning in sentences, and sentences obtain meaning in their contexts of use. At the science/society interface, words can be expected to have different meanings: the codes of communication that provide meaning to words differ on the varying sides of the interface. Furthermore, meanings and interfaces may change over time. Given this structuring of meaning across interfaces and over time, we distinguish between metaphors and diaphors as reflexive mechanisms that facilitate the translation between contexts. Our empirical focus is on three recent scientific controversies: Monarch butterflies, Frankenfoods, and stem-cell therapies. This study explores new avenues that relate the study of co-word analysis in context with the sociological quest for the analysis and processing of meaning.

**Keywords.** Codification; Translation; Mapping; Metaphor; Co-words; Interfaces

[i] University of Amsterdam, Amsterdam School of Communications Research (ASCoR), Kloveniersburgwal 48, 1012CX Amsterdam. E-mail. loet@leydesdorff.net; http://www.leydesdorff.net
[ii] Royal Netherlands Academy of Arts and Sciences, Virtual Knowledge Studio, PB 95110 1090 HC Amsterdam; E-mail. iina.hellsten@vks.knaw.nl

## Introduction

A large number of texts can be retrieved from the Internet for research purposes through the use of search engines, citation index databases, on-line archives of newspapers, scientific journals, popular scientific magazines, and on-line discussion groups; the websites and databases of various governmental, non-governmental, and commercial organizations can also be mined. This overload of textual materials poses new methodological challenges for the disciplines in the humanities and social sciences that are interested in text analysis. How can one automate the analysis of large amounts of texts that can no longer be analyzed qualitatively or coded manually, and still obtain conceptually meaningful and valid results?

Several research traditions, such as computer-aided content analysis, corpus-based linguistics, and the so-called 'sociology of translation' (Callon *et al.*, 1986; Stegman & Grohmann, 2003) have developed tools for the automated analysis of texts. Despite the different disciplinary backgrounds and research agendas of these traditions, they have all faced similar problems with the ambiguity of language. Words and the relations among words ('co-words') mean different things in other contexts, and the meaning of words can be expected to change, particularly in science where novelty production is part of the mission of the enterprise (Whitley, 1984; Fry, 2006). Without further reflection, words and co-words cannot be used for mapping the dynamics of science and technology (Leydesdorff, 1992; 1997).

In other words, one needs to specify a next-order mechanism of meaning exchanges to study both the changing distributions of words and the variance in their meanings and relations. For example, Luhmann (1984, 1986) argued that social systems communicate by processing meaning on top of information exchanges. From this perspective, meaning-processing is considered as a property of the systems of coordination in society. Meaning is generated by *positioning* the communication within networks of relations. Thus, meanings can be expected to vary across domains of use (e.g. science, journalism, economics). For example, one can generate value in economic transactions, but scientific theories are improved through discursive arguments. While Luhmann focused on the differentiation of meaning-processing and was not so much interested in the relation between meaning-processing and information-processing, this interface is precisely the challenge for the information scientist. We are interested in whether meanings can be traced and measured in the communications that occur between the different domains of use (such as the sciences, the economy, and the mass media) of society, and whether such mappings can be automated.

In computer-aided content analysis (e.g. Klein, 2004) the main focus has been on processing large bodies of textual data and on automatically coding specific aspects of the texts. Searching for particular words in documents, creating word frequency lists, and listings of word concordances have been automated, but within this tradition the coding schemes have to be developed by the analysts. As Krippendorff (1980/2002) notes, there remains a need to specify the context in which the texts become meaningful. Similarly, the main aim of research in corpus-based linguistics (e.g., Kennedy, 1998) has been to automate corpus analysis by tagging words within their grammatical contexts and clustering various tokens of the same word (e.g. 'word' and 'Words') as belonging to the same type. Yet, the problem of the semantic ambiguity of words has remained.

In science and technology studies, co-occurrences of words ('co-words') have been considered as the carriers of meaning across different domains (Callon *et al*., 1983). In the so-called 'sociology of translation' (Callon *et al*., 1986; Callon *et al*., 1991), co-words have been used to map the dynamics of science and technology in terms of translations. The main focus in quantitative studies of translations has been on the network of co-occurring key words as indicators of activity in the document sets (Callon *et al*., 1991; Ruiz-Baños *et al*., 1999; Stegmann & Grohmann, 2003; Bailón-Moreno *et al*., 2005). In the network of co-words, however, "the robustness of structured relations does not depend on qualities inherent to those relations but on the network of associations that form its context" (Teil & Latour, 1995).

Stegmann & Grohmann (2003) emphasized that co-words are particularly suited for the study of 'weak links' (Granovetter, 1973): the co-words relate otherwise unconnected literatures. These authors proposed to call this activity 'Swanson Linking' because in a series of articles Swanson (e.g., 1990, 1999) used this linking for discovering new relations like adverse drugs reactions (Rikken, 1998; cf. Rikken *et al.,* 1995). Our approach differs from these studies in that our focus is not on the relations and co-occurrences of words, but on the positions of words in different semantic fields. These *positions* can be considered as the unintended results of a set of relations in a network among agents or documents (Burt, 1982, 1983).

In other words, we are not only interested in dyadic co-occurrences, but also in single occurrences and triadic (etc.) co-occurrences. Accordingly, we will not use the co-occurrence matrix but the underlying asymmetrical matrix of documents versus words, and subsequently compute the distance among the word vectors using the vector-space model, that is, using the cosine as a similarity measure. The co-occurrence matrix—which contains less information—can be obtained by multiplying the asymmetrical matrix with its transposed (Leydesdorff, 1989; Leydesdorff & Vaughan, forthcoming).

Our specific focus is on science communication because at the interface between science and other domains of society, words can be expected to have different meanings. These domains use different codes for the communication, and also the degree of codification may differ across the domain of use. For example, in daily life, a 'shortage of energy' means something very different from the concept of 'energy' as a conserved quantity in physics. The degree of codification of the words is higher in scientific articles than in the mass media. Furthermore, in the sciences, meanings can be expected to change with the development of new knowledge.

As case studies, we use three scientific controversies that have flourished recently in public debates: first, Monarch butterflies; second, Frankenfoods; and third, stem cells. However, before turning to these case studies, let us first discuss in more detail the problem of automating the mapping of the meanings of the words and the question of what could be considered as providing the contexts for such mapping.

## Mapping translations between contexts: metaphors and diaphors

Information is codified when provided with meaning. Some meanings, more than others, gain resonance between the different domains in society. In the analysis of how meaning is given to the uncertainty contained in a distribution of words, one can distinguish between a diachronic problem and a synchronic problem. The synchronic problem is further complicated when different meanings—which can each be codified in different domains—are exchanged as in social systems. The synchronic and the diachronic mechanisms may further interact in a non-linear mode; meanings can then be stabilized locally and sometimes further be meta-stabilized and globalised, as in scientific communication.

Historically, the measurement of meaning has had two relatively independent roots. On the one hand, researchers have attempted to measure meaning from a psychological perspective using scales (Mitroff, 1974; Osgood *et al.*, 1957). On the other hand, information science research has focused on how the measurement of meaning can be operationalised using words and their co-occurrences. For example, in information retrieval, Salton and McGill (1983) proposed the cosine between word vectors as providing a spatial representation of how words are positioned in relations among words. Thus, the information-theoretical tradition is dominated by a semantic orientation on meaning as a structural property of the communication networks, while the psychological measurement can be considered as based primarily on a pragmatic theory of meaning. Our focus in this study is on measuring meaning within the semantic tradition, that is, as a property of the network of words.

In order to specify the context for measuring meanings, we focused on specific kinds of words, notably, words that can be considered as flagships for the debates. Specific terms such as "Frankenfoods" and "stem cells" are used in scientific and popular-scientific domains as well as in journalism, and may therefore provide common ground for the different discourses while still functioning differently in each of these domains. Maasen and Weingart (1995) discussed such words as metaphors which can be considered as 'messengers of meaning.' Metaphors would generate the dynamics of knowledge (Maasen and Weingart, 2000).

Our expectation is that a metaphor can be considered as one reflexive mechanism in the networks of words among others. A metaphor can act as a 'messenger of meaning' or a 'translator spokesman' in a symbolic manner because its occurrence is punctuated. Translation, however, can also be sub-symbolic, that is, a result of the interactions among different densities in the network. Translation in science communication may thus function both symbolically (as metaphors) and sub-symbolically (as diaphors). We hypothesize that metaphors and diaphors can be considered as tools of intermediation that channel meanings across different arenas in the communication of science; this is possible because they both contribute to carrying a set of relations from one domain to another.

A metaphor provides a mapping across two or more experiential domains (Lakoff and Johnson, 1980; Fauconnier and Turner, 2002). In the metaphor of "Frankenfoods," for example, food is perceived and experienced through the myth of Frankenstein's monster. However, the metaphor provides a perspective or window on issues and thus restricts the complexity in the system of reference. Metaphorical mapping from a source domain (in this case, the Frankenstein myth) to the target domain (GM foods) is always partial, since only some of the meanings associated with the source domain are evoked. This depends on the context in which the metaphor is used.[1] Condit *et al.* (2002), for example, compared the metaphor of 'genes as recipes' to that of 'genes as blueprints,' and found that "meaning depends on selections from a polysemic universe of associations of metaphoric vehicle and the polyvalent responses to each of these associations."

The concept of 'diaphor' was suggested by Luhmann (1990) in order to make an analytical distinction between words that carry meaning (i.e., metaphors), and words that contribute to the boundary construction between domains of communication in discourses (Weelwright, 1962). Whereas metaphors such as "Frankenfoods" can be considered as punctuated tools of intermediation that channel meanings among otherwise different semantic fields, common words such as "stem cells" obtain meaning from their positions in the field of relating words. A metaphor brings domains together in a symbolic mode, while common words are expected to function sub-symbolically; their contribution to the translation of meaning is the result of interactions among the various clusters and hubs in the networks of words on the different sides of an interface. In the sub-symbolic case, the tensions found in the meaning of these terms are not necessarily resolved.

Both diaphors and metaphors can be studied diachronically and/or synchronically. In this study, we limit the analysis to a diachronic discussion of the metaphor and a synchronic comparison in the case where we expect a diaphor. Thus, we focus on the two extreme poles of a continuum of potentially different mechanisms of codification. A very pronounced metaphor ("Frankenfoods") is studied in a largely un-codified set of documents, and a common word ("stem cell") in a set of codified texts. However, we first validate our methodology by using a qualitative study of five documents central to the controversy about the potentially harmful effects of genetically modified corn pollen on Monarch butterflies. This case allows us to build upon an argument made by Nucci (2004) and McInerney *et al.* (2004) that in the translation of science to various publics, the frames of reference are different in the various domains and their related discourses. Can the differences in meaning indicated by these authors be made automatically visible by using our methods?

[1] Different types of metaphors may function differently. The metaphor of Frankenfood is a discourse metaphor (Zinken *et al.*, forthcoming) that has been used as a 'one-issue' metaphor to oppose GM foods. More general metaphors, such as 'politics as game,' can occur in a variety of linguistic expressions and may therefore function differently.

**Methodology**

The datasets will be specified below in the three case studies separately, but we utilised a common methodology in all three cases in order to reduce the complexity in the comparison. After the first case study that—as noted—focuses on the debate about the genetic modified corn pollens and the Monarch butterfly, we scale up in a second step to sets of documents that can no longer be read and coded manually. Here, we draw upon two previous case studies in which we have developed techniques to trace mechanisms for reflection among textual domains. In one study, Hellsten (2003) traced the metaphor of "Frankenfoods" on the web over time. In the other, Leydesdorff & Hellsten (2005) used the diaphor "stem cells" to map words and co-words in contexts across different domains like newspapers, the Internet, and scientific databases.

Our techniques are based on commonly available software programs. The document sets were downloaded from the Internet and saved in the .html format. When we downloaded large data sets (as in the second and the third case study), we used the Internet module available in Visual Basic. In all case studies, the files were first parsed so that each document represented a separate text file. These documents were then broken down into sentences and words. Word frequency lists were generated and we used the stop word list of the U.S. Patent and Trade Office (located at http://www.uspto.gov/patft/help/stopword.htm) throughout the study.[2] Furthermore, the plural *s* was always removed. We selected only the body text for analysis—in some cases the full text, and in the case of large sets the titles—thus excluding additional information included on the web pages, such as 'print the document' icons and other elements that are not part of the actual text.

[2] The stop word list of the USPTO is designed in the context of specialised patent jargon, but we use it also for other types of documents for reasons of consistency. Our main aim here is to introduce a new method. Of course, there is a large number of other stop word lists available or one could also use statistical approaches (Bookstein *et al.*, 1995).

Dedicated software was written that uses these texts and outputs asymmetrical matrices with textual units (titles, paragraphs, or documents) as the cases and words as variables.[3] These matrices can be imported into Excel, SPSS, and UCINET for statistical analysis. UCINET files can be exported to the visualization program Pajek.[4] Semantic maps were drawn on the basis of the similarities among word distributions using the cosine for normalization (Ahlgren *et al*., 2003; Hamers *et al*., 1989; Salton and McGill, 1983).[5] The maps were optimized for visualization using pragmatic cut-off levels of word frequencies in order to keep them readable. We used approximately one hundred words as the maximum. (Technically, it is possible to include many more words in the analysis, but then the reading of the maps becomes problematic; this issue will be addressed below.) The visualizations are based on the algorithm of Kamada and Kawai (1989) as it is available in Pajek.[6] Factor analysis was conducted in all of the case studies as an exploration of the main dimensions in the networks of words.

## Results

### *The case of the Monarch butterflies*

The debate about Monarch butterflies began in May 1999 when *Nature* published Losey *et al*.'s Scientific Correspondence (Losey *et al*., 1999). The letter discussed the preliminary

---

[3] This routine (fulltext.exe) can be retrieved at http://www.leydesdorff.net/software/fulltext .
[4] The homepage of Pajek can be found at http://vlado.fmf.uni-lj.si/pub/networks/pajek/
[5] Salton's cosine is defined as the cosine of the angle enclosed between two vectors *x* and *y* as follows:

$$\text{Cosine}(x,y) = \frac{\sum_{i=1}^{n} x_i y_i}{\sqrt{\sum_{i=1}^{n} x_i^2}\sqrt{\sum_{i=1}^{n} y_i^2}} = \frac{\sum_{i=1}^{n} x_i y_i}{\sqrt{(\sum_{i=1}^{n} x_i^2)*(\sum_{i=1}^{n} y_i^2)}}$$

This formula for the cosine is very similar to the one for the Pearson correlation coefficient except that the latter measure normalizes the values of the variables with reference to the mean (Jones & Furnas, 1987; Ahlgren *et al*., 2003).
[6] This algorithm represents the network as a system of springs with relaxed lengths proportional to the edge length. Nodes are iteratively repositioned to minimize the overall 'energy' of the spring system using a steepest descent procedure. The procedure is analogous to some forms of non-metric multi-dimensional scaling. A disadvantage of this model is that unconnected nodes may remain randomly positioned across the visualization. Unconnected nodes are therefore not included in the visualizations below.

results that showed the potentially harmful effects of the pollen of genetically modified corn on Monarch butterfly larvae. Cornell University—the institution of these researchers—immediately published a press release on the results. The Union of Concerned Scientists (UCS) and Greenpeace reacted to the topic, and published their own press releases. Finally, the Biotechnology Industry Organization (BIO) tried to counter these press releases by issuing one of its own, in August 1999 (Annex 1).

Nucci (2004) analyzed the different rhetoric used in these five documents to illustrate how scientific information is carried across media boundaries. She states that "rhetorical changes altered the story and most likely served as a catalyst for the media frenzy that accompanied the article." In order to test our methodology, we first show that by measuring the meanings of the (co-)words in these five documents, we are able to visualize the rhetorical changes, that is, the different frames indicated by Nucci.

From our methodological perspective, the paragraphs in the five documents provide us with the cases to which the words are attributed as variables. Only eight (non-stop word) words are used in all five documents ('pollen,' 'corn,' 'monarch,' 'field,' 'butterfly,' 'feed,' 'grew,' and 'laboratory') and only two of these words, namely 'pollen' and 'monarch,' occur more than twice in each of the documents. We focus on these two words in order to show the change of the positions. In order to sort out how these words are positioned in the different documents, we will draw semantic maps using all the words that occur at least twice in a given document.

As these are single document studies, the cosine threshold for inclusion in the graph is set at the level of larger than or equal to 0.5 (Chen, 2003). The cosine values are affected by the density of the relations: the tighter the network, the higher the threshold has to be set in order to produce a map that exhibits the semantic organization. Unlike document sets, single documents provide 'restricted discourses' that one can expect to be well organized in word usage and tightly connected, while one can expect that 'elaborate discourses' among documents are more loosely organized (Bernstein, 1971; Coser, 1975; Leydesdorff, 1997). For this reason, we shall use a threshold of cosine $\geq$ 0.1 in the case of large document sets.

Let us first turn to the analysis of the original research paper published in *Nature*, and then proceed via the press release by the university to that of the UCS. After these university-based documents we analyse the press releases of Greenpeace, an organization that is against GM foods for more general reasons, and the subsequent reaction by BIO, an organization that lobbies in favour of using biotechnology.

In the research report published in *Nature,* 710 words were used in 8 paragraphs. Among the 234 unique words, the 59 words which occurred more than once were selected for the analysis. In the semantic map that results (Figure 1), the two words that were our focus, namely 'pollen' and 'Monarch,' are part of different word clusters, thus illustrating how they embody different parts of the argument. In order to draw attention to the clusters that we wished to focus on, we illustrate them with grey shades. The methodology of the research is visible as a third grouping. As expected in the case of scientific literature, the different parts of the argument are clearly separated from one another in terms of the cause, the effect of the problem, and the work process that validates the inference (Leydesdorff, 1991).

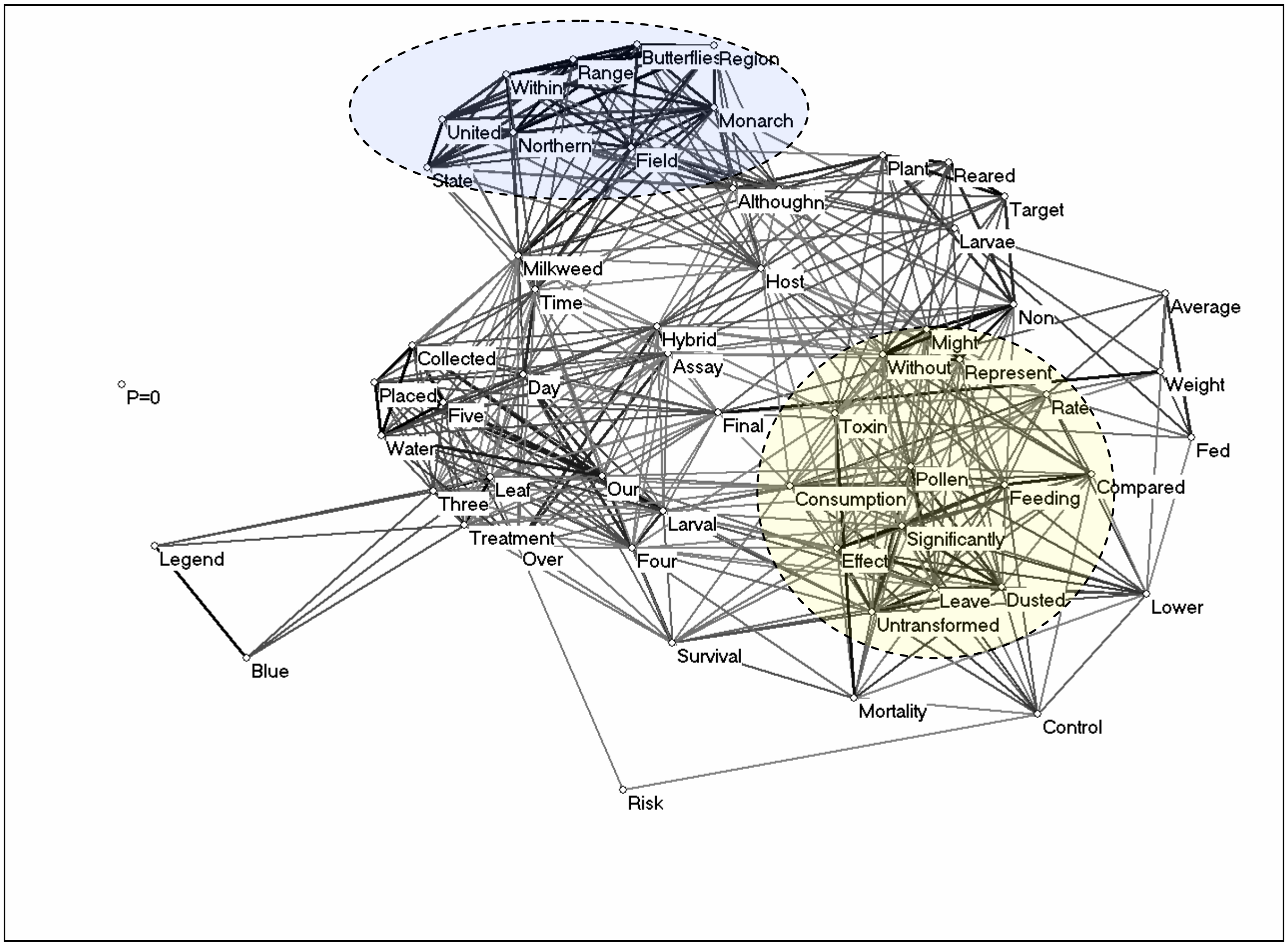

**Figure 1:** The cosine map of 59 words used more than once in the Scientific Correspondence published in *Nature*, 399: 214 on May 20, 1999 (cosine ≥ 0.5).

Unlike the practice in corpus-based linguistics, we did not group the tokens ‘larvae’ and ‘larval’ as a single type; in the figure they are grouped differently. In a six-factor solution of the matrix (which explains 94.2% of the variance), for example, ‘larval’ has a factor loading of 0.855 on factor two—mainly representing methodological words—while the word ‘larvae’ loads on the fifth factor with –0.758. (There is not a lot of interfactorial complexity in the orthogonally rotated solution.[7]) The ‘larvae’ are among the subjects of study, while the word ‘larval’ belongs to the methods section of the argument. These distinctions are very sensitive in scientific literature (Leydesdorff, 1997). If we had grouped these words together in a coding scheme *ex ante*, the semantic map would have been distorted.

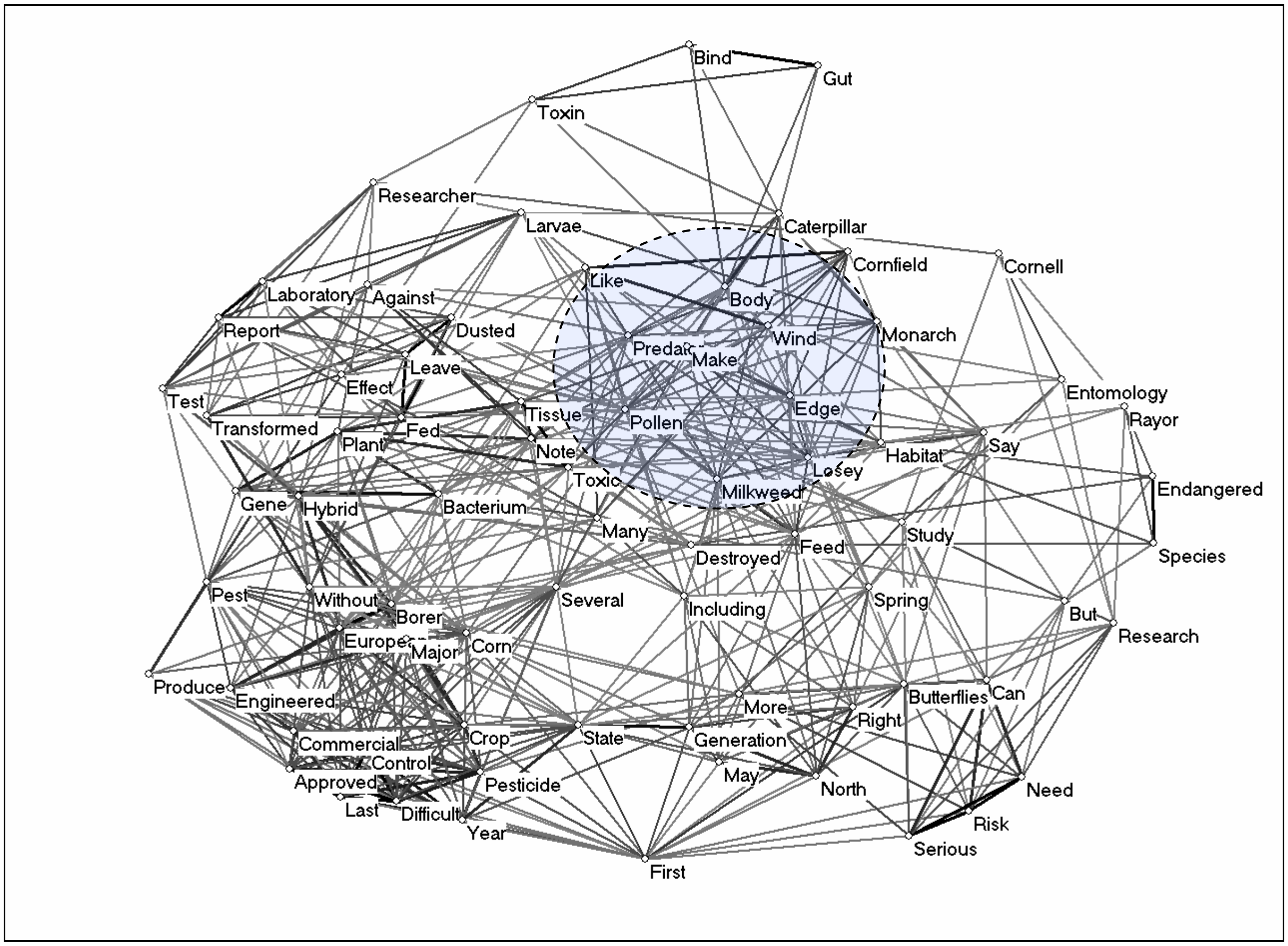

**Figure 2:** The cosine map of 77 words used more than once in the Cornell University press release released on May 19, 1999 (cosine ≥ 0.5).

[7] The loading of larval on Factor 5 is .228 and the loading of ‘larvae’ on Factor 2 is below 0.1.

The next picture (Figure 2) provides a similar representation of the press release by the home university of the research group. This press release of Cornell University consisted of 12 paragraphs and 795 words, of which 296 were unique. Seventy-seven words occurred more than once and were therefore included in the analysis. Unlike the article in *Nature*, the main common words, ‘pollen’ and ‘Monarch,’ are here part of the same word cluster. The argumentative structure of the scientific contribution is merged in this reflection with another purpose, notably to draw attention to the main findings of the researchers. The *Nature* article is reflected from an external angle in the press release, and the possible *implications* of the findings are emphasized.

Furthermore, this map shows that the press release raised a new topic that relates to the European corn borer—against which the corn was genetically modified. Whereas *Nature* talked about ‘larvae’, the press release uses both the terms ‘caterpillar’ and ‘larvae.’ ‘Caterpillar’ occurs in the word cluster with the words ‘pollen’ and ‘Monarch’ whereas the word ‘larvae’ is oriented towards a separate cluster with words like ‘laboratory’ and ‘report,’ that is, when referring to the research process. The science communication induces this distinction between the scientific word and the more common word usage.

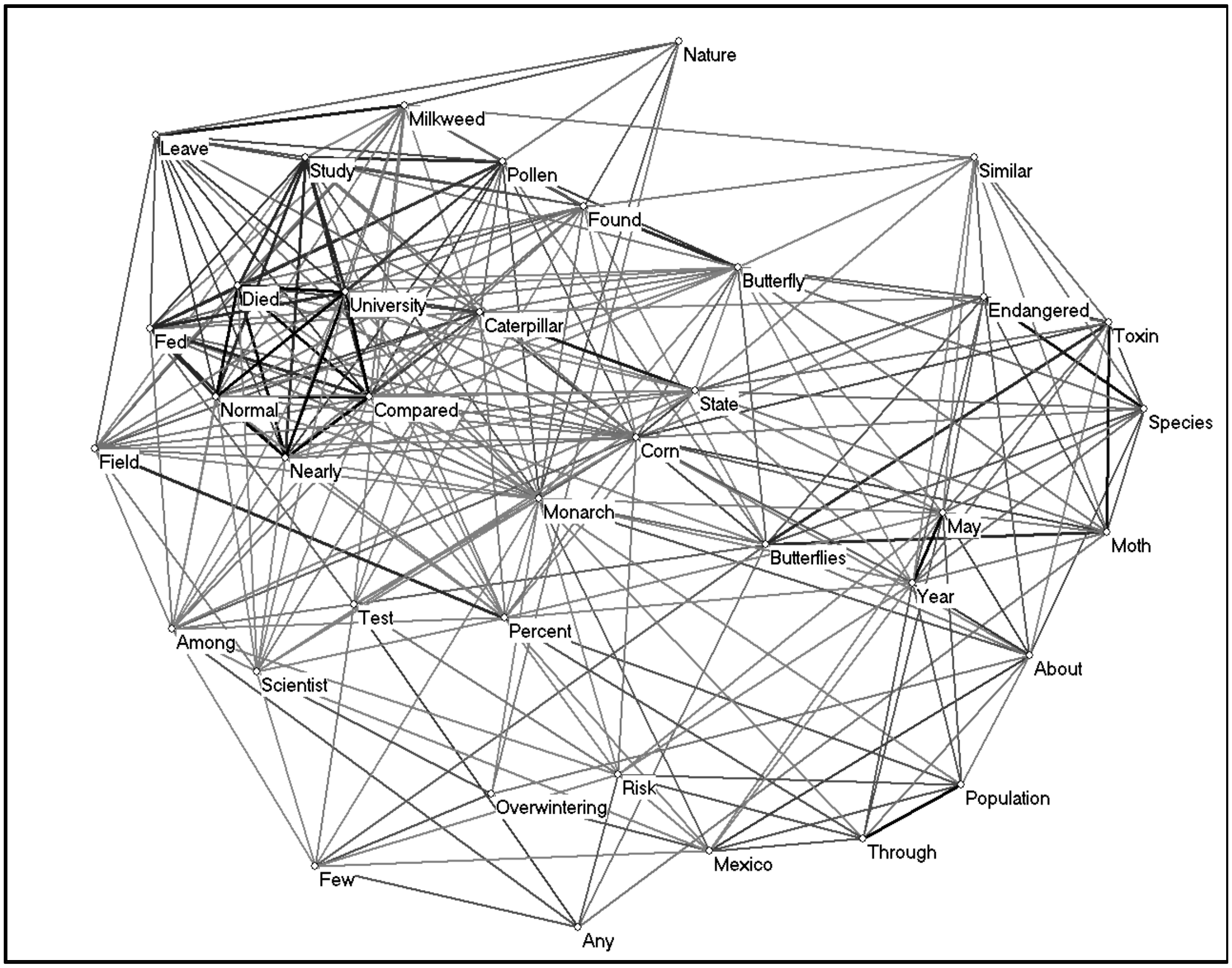

**Figure 3:** The cosine map of 38 words used more than once in the UCS document published in the May, 1999 issue (cosine ≥ 0.5).

We expected that in the press release by the *Union of Concerned Scientists* (Figure 3), the words 'pollen' and 'Monarch' might again be presented in separate word clusters because this press release built directly upon the original letter in *Nature*. However, this was not the case. The UCS press release contains 7 paragraphs and 454 words. Only 38 words occurred more than once, and therefore form the basis for the semantic map. In Figure 3, the words 'Monarch' and 'pollen' appear as parts of the same component, although a bit more separated than in the university press release. In this document, the word 'Monarch' holds a central position. The frame has thus shifted from the genetically modified 'pollen' (the cause) to the Monarch butterfly as an endangered species (the consequence). The word 'larvae' is not used, and the term 'caterpillar' is part of the same word cluster as the words 'pollen' and 'Monarch.' The argument is mainly popularized.

Let us next take a look at the documents of Greenpeace and BIO. The arguments made in these releases can be expected to differ from each other since the former organization opposes GM foods and the latter lobbies for biotechnology in general. The document of Greenpeace consists of 7 paragraphs and a total of 442 words. We selected the 38 words that occurred more than once for the visualization (Figure 4).

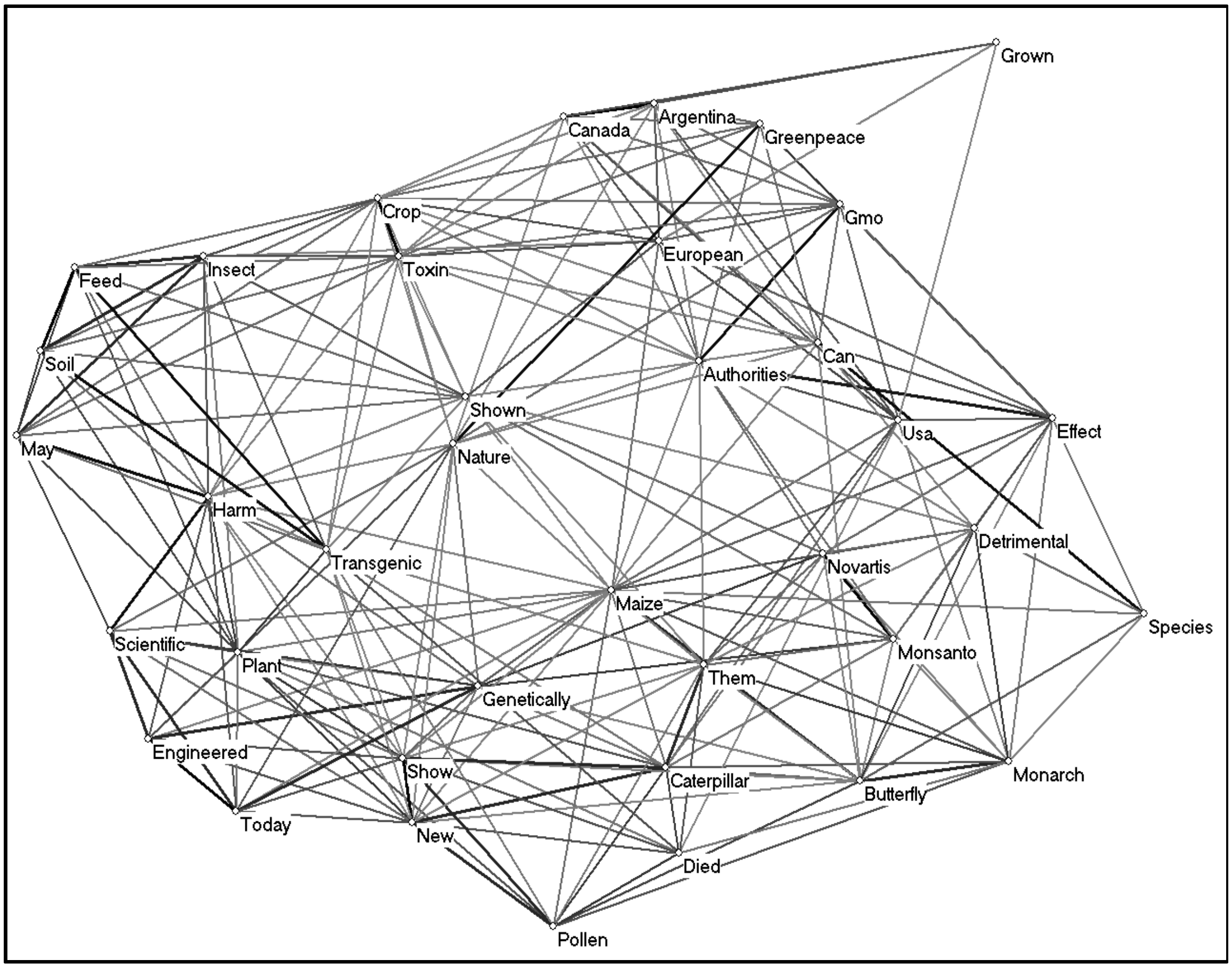

**Figure 4:** The cosine map of 38 words used more than once in the Greenpeace document published on May 20, 1999 (cosine ≥ 0.5).

In Figure 4, the words 'pollen' and 'Monarch' again belong to the same word cluster. The cluster is at the margin of the figure because the main concern is not with the discovery, but with its social consequences. Further, the word 'caterpillar' is part of the same cluster of words including 'pollen', 'died', and 'Monarch,' as in the Cornell University press release. However, the word ‘larvae’ is not used. The words ‘Nature’ and ‘maize’ hold central positions in the map. In addition, we interpret the map to show an orientation towards other countries, such as Canada and Argentina, as well as the European continent, and towards the biotech companies behind the development of the corn in question.

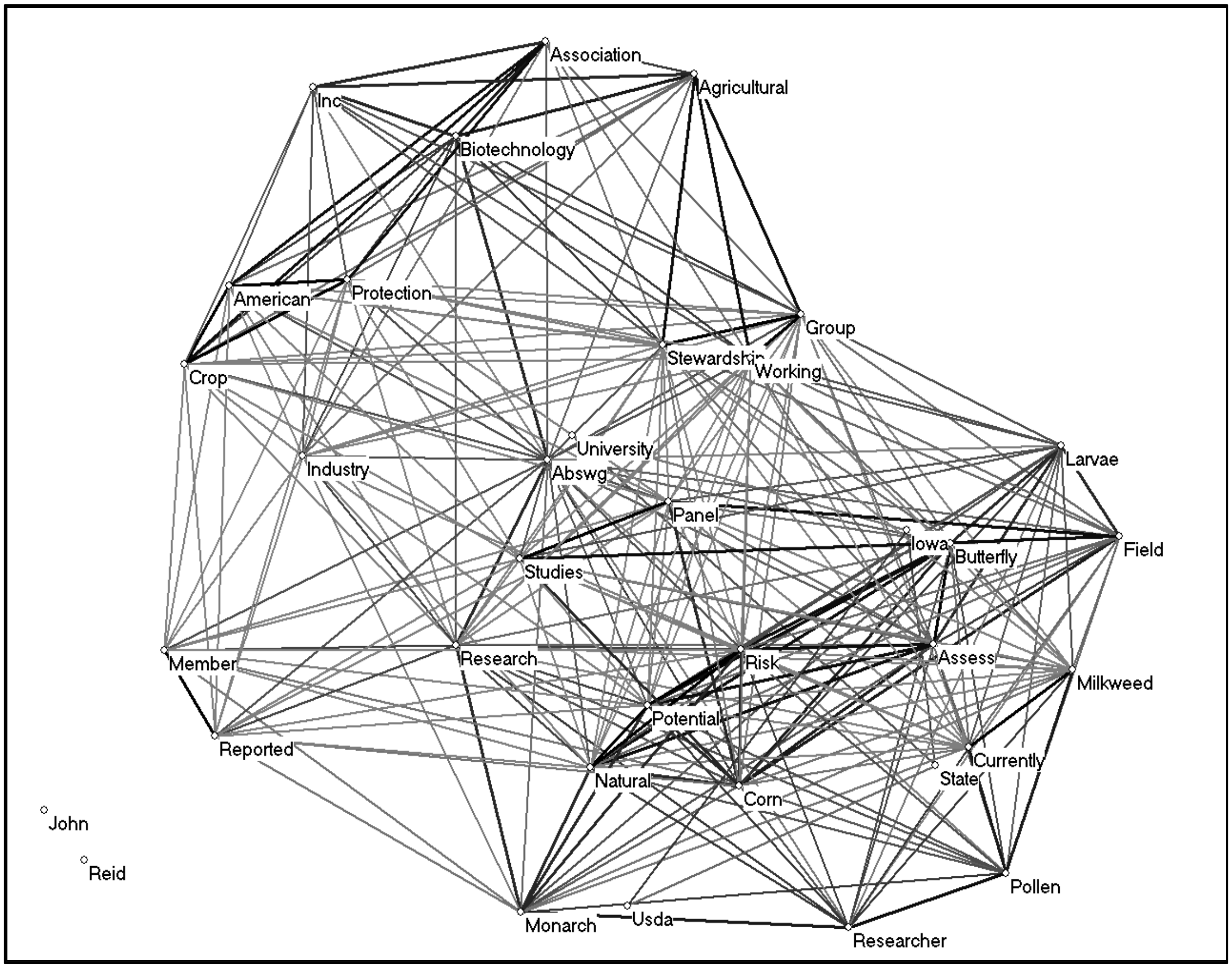

**Figure 5:** The cosine map of 36 words used more than once in the BIO document, published on 12 August, 1999 (cosine ≥ 0.5).

In the BIO document the words ‘pollen’ and ‘Monarch’ are also part of the same word cluster (Figure 5). The document consists of 6 paragraphs that contain 361 words. Thirty-six words occurred more than once and were selected for the analysis. The word clusters are different from the map of Greenpeace in that the emphasis is on the ‘potential risks’ instead of scientific research that ‘shows’ the risks. The word ‘larvae’ is used instead of ‘caterpillar.’ In the map, there are two unique word clusters: one centres around the agricultural biotechnology association and the other focuses on the protection of American crops and industry.

In conclusion, we were able automatically to filter out semantic differences between these five documents. As suggested by Nucci (2004), the frames of the documents were different. This could be analysed and visualized using the network of co-occurring words. However, our analysis remains purely semantic. One cannot indicate the rhetorical value of the claims without reading the documents, or without content analysis, because these pragmatic elements belong to another dimension of the communication.

The technique enabled us to detect that the main change in the semantics of the co-words occurred when the topic moved from the scientific context of *Nature* to the various press releases, including the press release by the university. The expectation of audiences seems to guide the selection of the frames of reference. In the semantic maps, one can also see novel topics across the various domains, such as the focus by the UCS on the butterfly instead of the pollen. While these five documents can also be coded manually, our purpose was to develop these techniques for larger document sets; the following two case studies use large sets of texts as data.

*The dynamics of frames: the case of Frankenfoods*

The metaphor of Frankenfoods, coined in 1992, gained popularity after 1998 when NGOs such as Organic Consumers (www.organicconsumers.org) and Genetically Manipulated Food News (home.intekom.com) in the U.S. and Friends of the Earth in Europe, mainly in the UK (www.foe.co.uk), began using it in calling for consumer action against GM foods. The metaphor was taken up by the newspapers and also generated discussion in Usenet (open-access) newsgroups.

The metaphor has been used in a wide variety of contexts on the Internet. Some of the web pages, for example, suggest that Frankenfoods are good for Halloween parties. Others report on the launch of a new *Oxford Dictionary of Phrase and Fable* in 2000 where "Frankenfoods" was one of the new entries. By 2001, the metaphor faded from the public agenda. In a previous study using qualitative methods, Hellsten (2003) showed that the contexts in which this metaphor was used changed from consumer concerns (NGOs) and the subsequent reaction by industry (*Monsanto*) to an issue on the political agenda (*The Times*).

The metaphor served different functions in these different domains of use. For the NGOs, the metaphor was useful in sparking emotions that can be transformed into action against genetic manipulation in food production. For the participants in the newsgroups, the metaphor effectively gave a name to these concerns. In newspapers, it provided a catchy and concise way of talking about the politicized issue.

In order to map the largest possible variety of contexts in which the metaphor was used over time, we turned to the Internet as a 'common' domain and used results generated by the *AltaVista Advanced Search Engine* in for the years between 1996 and 2003. *AltaVista* has well-known shortcomings, such as search engine instability and updating of the web pages (Bar-Ilan, 2001; Rousseau, 1999; Thelwall, 2001). The *AltaVista Advanced Search Engine*, however, allows us to search for data by calendar year even if the web pages are constantly overwritten; the date indicates when the page in question was last modified and the crawler notices this change (Wouters *et al.* 2004). Despite these shortcomings, the data is suitable for illustrating our techniques.[8]

The summaries and titles of the documents were downloaded from the Internet on January 21 - 23, 2004, using the Boolean search string 'Frankenfood* OR (Frankenstein AND food*).' The analysis was limited to title words during the years 1996 to 2003, and semantic maps were drawn for each of these years. Word frequency lists were created and the cosines among the word vectors calculated. Here, we show only three of these years (1996, 1999, and 2003) because these three pictures allow us to make our point (which is based on both analyzing the in-between years and the factor-analytic results).

For the calendar year 1996, AltaVista reported 125 documents of which we were available to retrieve 74.[9] In the 74 titles of these documents, 233 different words were used, and 44 of these words occurred more than once. As the number of co-occurring words is below the pragmatic cut-off level (approximately one hundred words), all 44 words were included in the analysis. The cosine threshold was set at cosine ≥ 0.1 because the similarity among the distributions of words used in this unrestricted domain is expected to be low (Salton & McGill, 1983; Leydesdorff, 1989).

[8] The *AltaVista* search engine was completely restructured in April 2004 when it was merged with the search engine of *Yahoo!*.

[9] This strong reduction in the number of documents retrieved is a consequence of setting the site filter deliberately on. This reduces the repetition of the same pages from one web site.

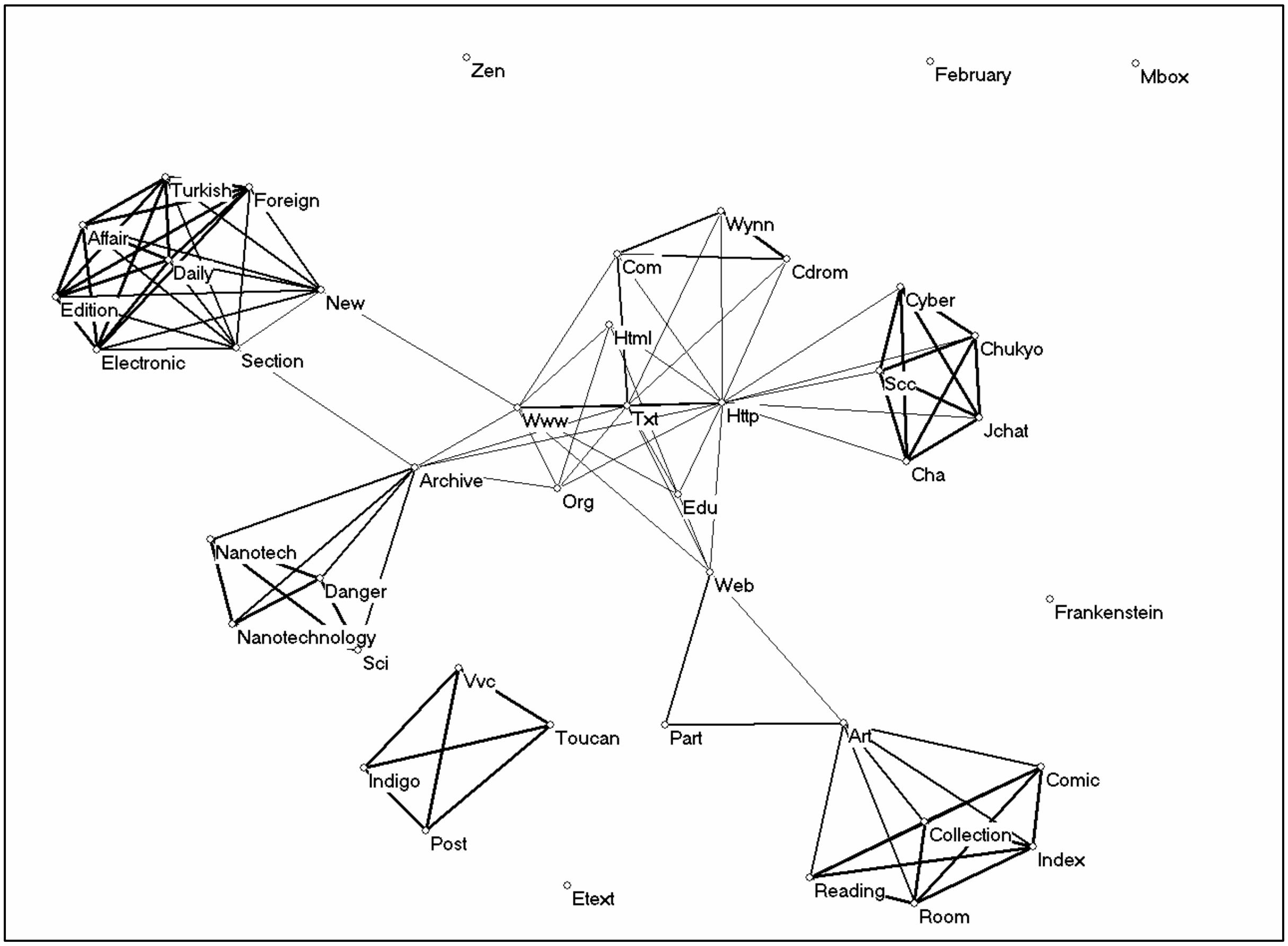

**Figure 6:** The cosine map of 44 words used more than once in the 74 documents on Frankenfoods in 1996 (cosine ≥ 0.1).

In this semantic map (Figure 6), there are a few clusters of words that reflect the debate in discussion forums and archives on the Web. In the titles of the documents, the metaphor of Frankenfood was not yet used in 1996, and even the word 'Frankenstein' is still unrelated to the word clusters. Frankenstein food was an emerging topic in the *AltaVista* domain of that year. As we used a list of stop words provided by the U.S. Patent Database for reasons of consistency, some of the most commonly co-occurring words on at the web like http, www, org, and edu were not suppressed. These words play a central role in the map in this relatively small set of title words. The other main clusters of words are around the dangers of nanotechnology, and news clippings published in the *Turkish Daily News – Electronic Edition*. In summary, the metaphor was not yet established on the Internet at that time.

For the year 1999, the *AltaVista* reported 957 documents of which 205 could be downloaded. Using the same threshold as above, the 105 words occurring more than once were included in the semantic map (Figure 7).

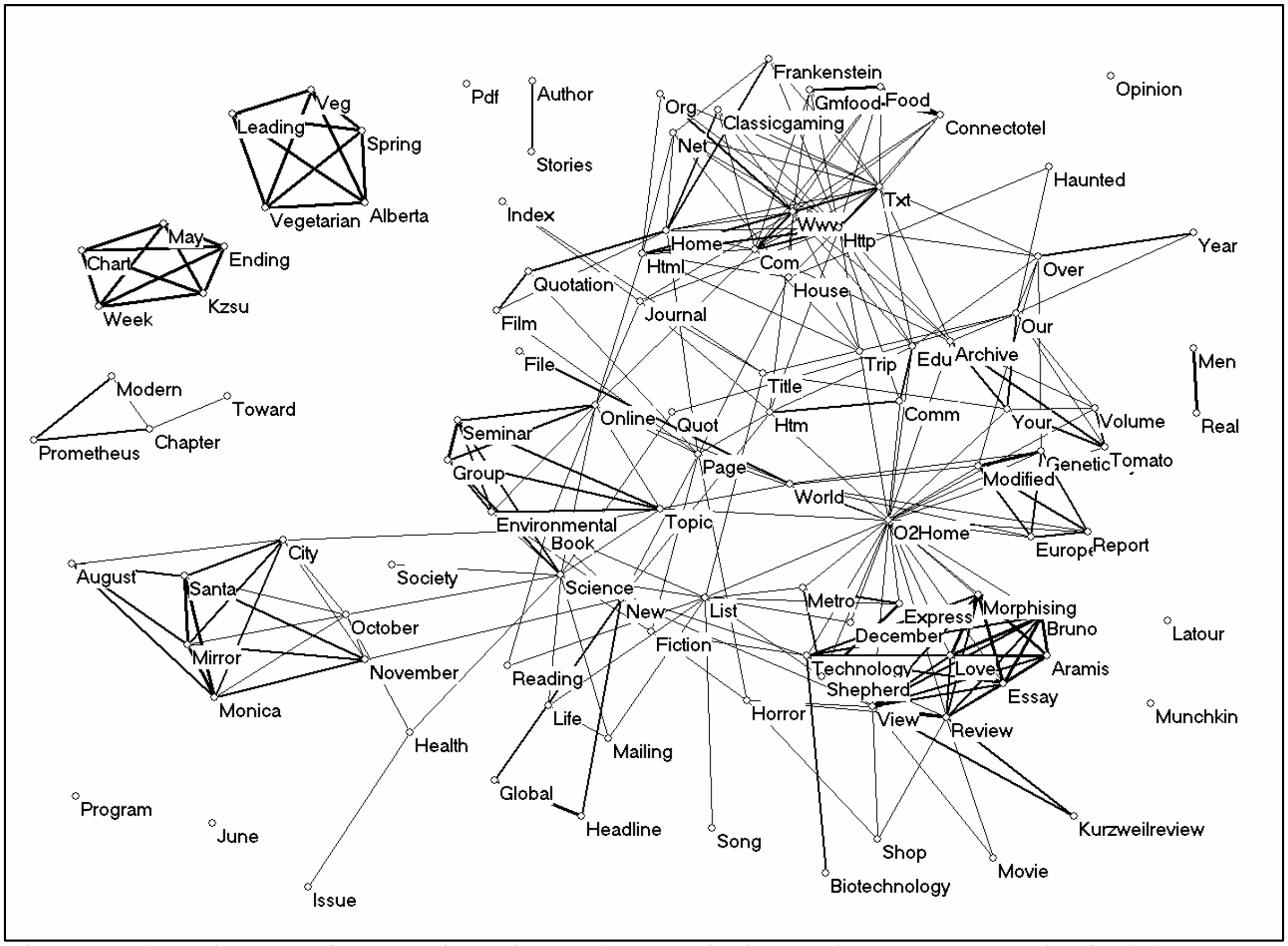

**Figure 7:** The cosine map of 107 words used more than once in the 205 documents on Frankenfoods in 1999 (cosine $\geq 0.1$).

The network of words in 1999 shows clearly delineated word clusters. The main words cluster around reviews of Bruno Latour's book (1996) *Aramis, or the Love of Technology*, and a report on a genetically modified tomato. A separate cluster of words on the left side of the picture represents the *Santa Monica Mirror*'s television programmes that are published on-line. The terms http, www, etc., no longer hold a central position in the semantic map, partly because of the higher number of meaningful documents and words included in the analysis. (This reflects that fewer documents than in 1996, used their URLs also as titles.) The network of websites using the metaphor of Frankenfoods and/or Frankenstein in combination with the word ‘food’ in their titles has been expanded in comparison to the year 1996 and became organized in relation to the relevant contexts.

For the year 2003, *AltaVista* reported about 45,336 documents using the search term. Only 6,101 of the documents were actually retrievable. These documents contained 8,235 different words that occurred 25,985 times. By using a threshold of 31 or more occurrences of the words, the number of words included in the analysis was reduced to 100 (Figure 8). The semantic map for the year 2003 shows a dispersed net of co-occurring words. The network no longer has a centre or a cluster structure. This may partly be due to the high threshold of 31 or more occurrences of the words which was needed in order to keep the map readable. However, it is unlikely that the less frequently occurring words could tie this network together. ‘Food’, for example, occurs 397 times in this set. The word cluster on ‘accelerated quiz list’ is a prominent but isolated cluster of words. Another cluster shows words such as ‘organic,’ ‘food,’ ‘genetically,’ ‘modified,’ and ‘engineered,’ but the meaningful clusters are no longer related.

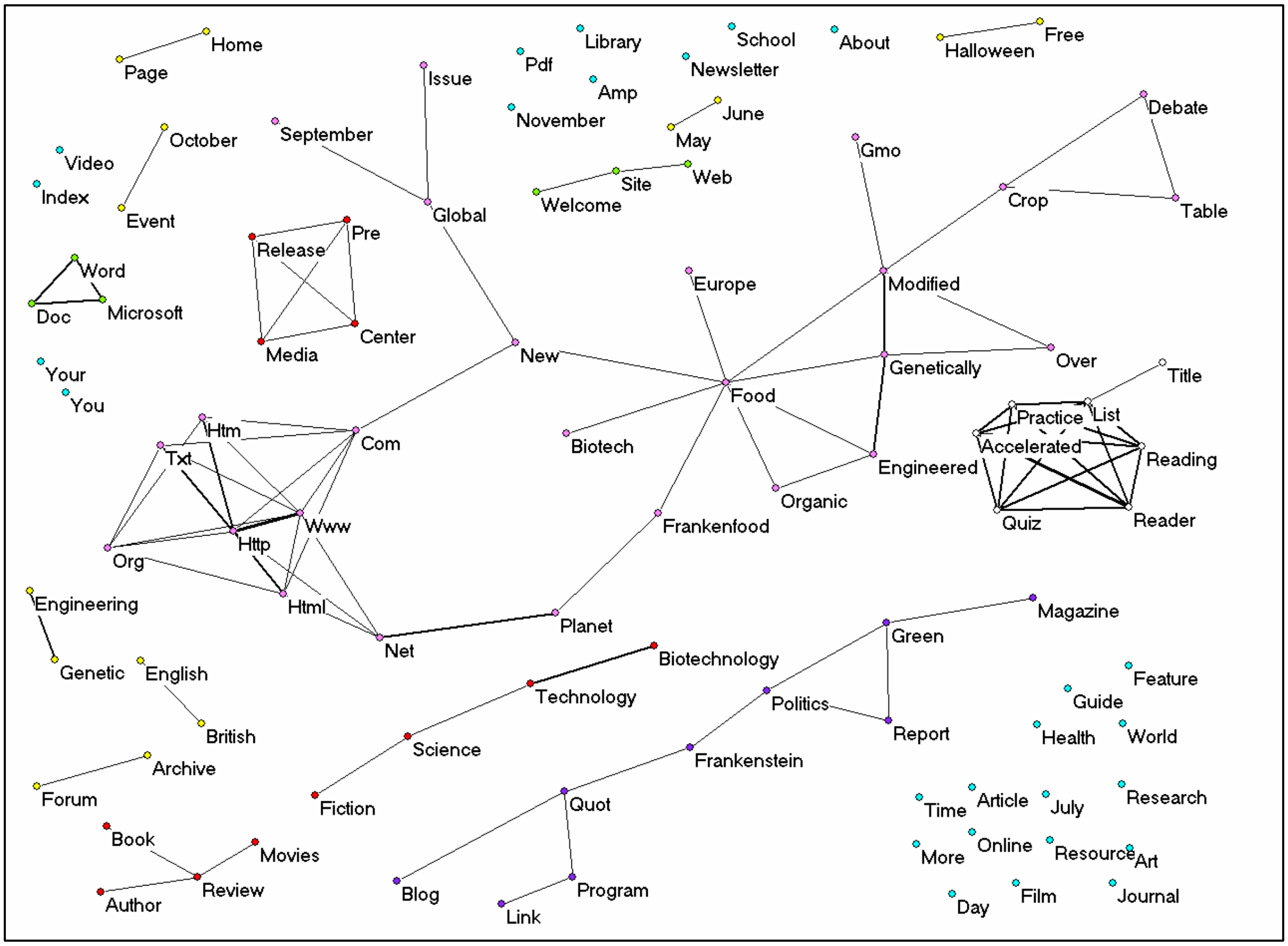

**Figure 8:** The cosine map of 100 words used more than 31 times in the 6101 documents on Frankenfoods in 2003 (cosine $\geq$ 0.1).

Our interpretation of these results is as follows: the decline of the organizing power of the metaphor was rapid in 1999 and 2000 when the metaphors of 'Frankenfood' and 'Frankenstein food' began to be outdated. Due to its generalized meaning, the metaphor was used increasingly across domains and therefore lost its domain-specificity and the ability to organize distinctions among domains. This might also explain why the NGOs stopped using the metaphor in 2000 (Hellsten, 2003). From this perspective the metaphor can be considered as an *anti-codifier*: the metaphor mediates meaning among contexts and thus blurs boundaries. The three figures presented above show the life cycle of the metaphor. The year 1999 provides the peak in the codification among co-occurring title words. Further research is needed to see whether other kinds of metaphors function similarly, that is, whether metaphors function as anti-codifiers over time in being used across boundaries.

The number of documents that form the basis of the analysis seems to affect the results: the more documents analysed, the more variation the semantic maps show. In other words, a single document is more codified than a set of documents. In the next section, we proceed from a single set of texts to a set of sets of texts, and explore how the differences among them affect the relative codification in the meanings of the words.

*Measuring the meanings of 'stem cells' across domains*

"Stem cells" have been an object of research since the 1960s. Progress in stem-cell research has been rapid from since 1990 and since the mid-1990s has provoked vivid public debate on the technical aspects involved. The advances in health care promised by this line of research, together with the ethical and social implications associated with stem-cell creation and exploitation in research, have attracted the attention of many groups, who perhaps not understanding the technical literature, often use the term differently in the relevant domains.

Our first map is based on the television address delivered by U.S. President George W. Bush on August 9, 2001 (Figure 9). This was the first time an American President had delivered a speech on national TV in a special broadcast focusing on a bio-ethical issue. He instructed the government-funded National Institutes of Health (NIH) to limit research funding to the 60 stem cell lines that the NIH had already recorded to date, and ordered the Institute not to create new lines (a process that requires using discarded human embryos). Many of the existing cell lines, however, proved unsafe for clinical trials because they had been grown on mouse media. This political decision thus interfered with an ongoing research process.

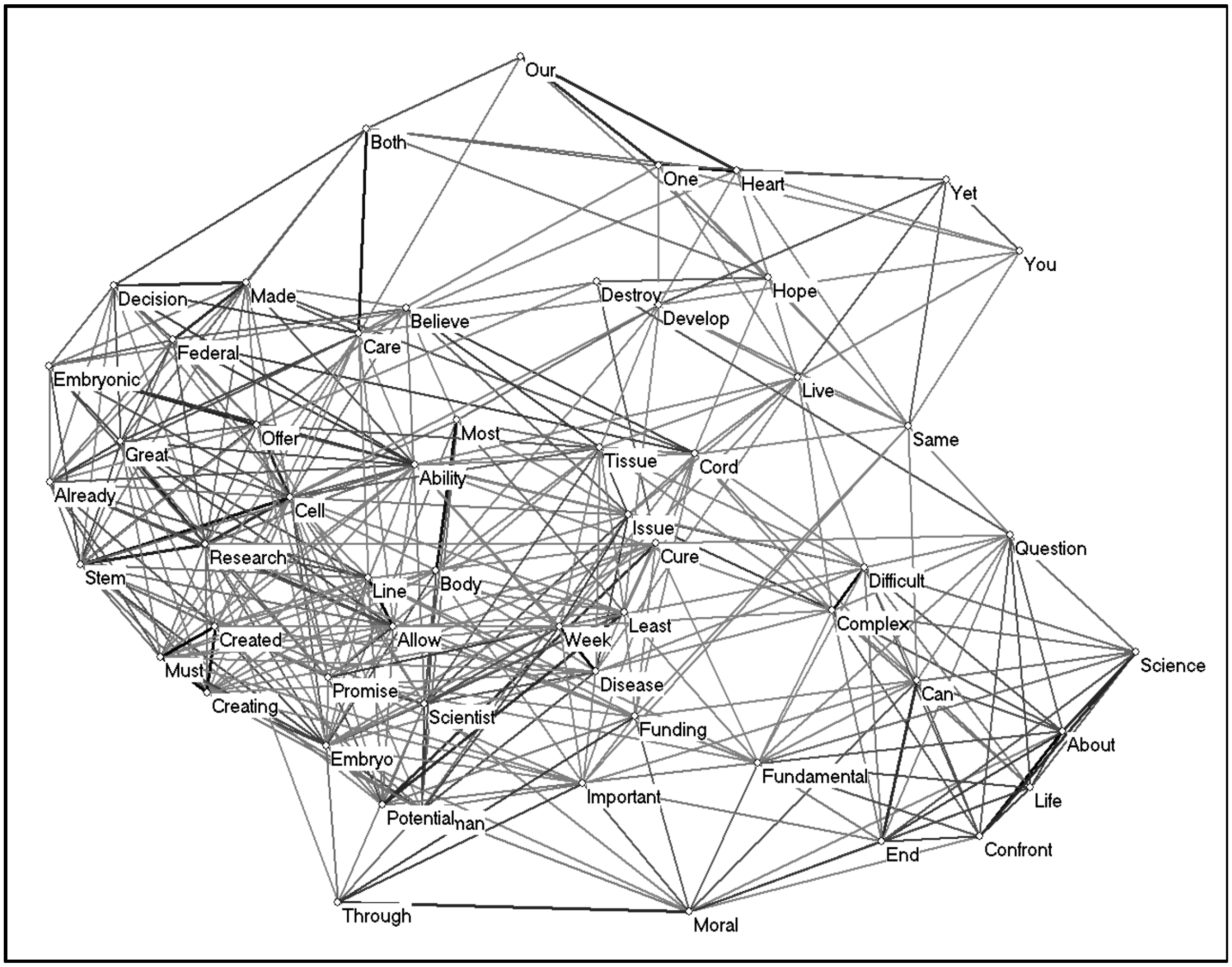

**Figure 9:** The cosine map of 57 words used more than once in President Bush's television address on stem cells of August 9, 2001 (cosine ≥ 0.5).

The stem cell debate further escalated in November 2001 when President Bush convinced the U.S. Congress to ban reproductive and therapeutic cloning, a ban that would directly affect the production of stem cells. Wertz (2002) noted that this ban does not extend to private-sector laboratories that do not receive government funds. Only therapeutically- oriented research funded by the government hadhas been banned.  Because of its diagnostic potential, stem-cell research thus became a subject of public controversy.

The text of the speech of President Bush's speech Bush on August 9, 2001, can be retrieved at http://www.whitehouse.gov/news/releases/2001/08/20010809-2.html. In this speech, 57 words were used more than once and are hence included in the analysis. Figure 9 visualizes the various clusters of words that the President used in his public appearance. The core of the argument is composed of a large number of words including "stem cell research" on the left side of the picture. Words like ‘our,’ ‘hope,’ and ‘heart’ are together in a looser cluster at the top. The words ‘fundamental,’ ‘difficult,’ ‘question,’ ‘confront,’ ‘science,’ and ‘life’ appear in a cluster on the lower right side as a separate grouping. As is often the case in single documents, the argument is well structured. (Consequently, we use the level of cosine ≥ 0.5 for the visualization.) There are eleven paragraphs, of which the last one consists only of the phrase ‘Thank you for listening.’

Second, to see how the issue was represented in a set of documents in the mass media, we collected articles that used the words “stem cell” or “stem cells” in the headlines of items published in *The New York Times* during 2001. Here, we are particularly interested in how the complex issue of stem cell research was popularized for wider audiences. The newspaper published 127 documents in 2001 where the topic was mentioned in the title. The semantic map (Figure 10) is based on 81 words that occurred more than once in the titles of these documents.

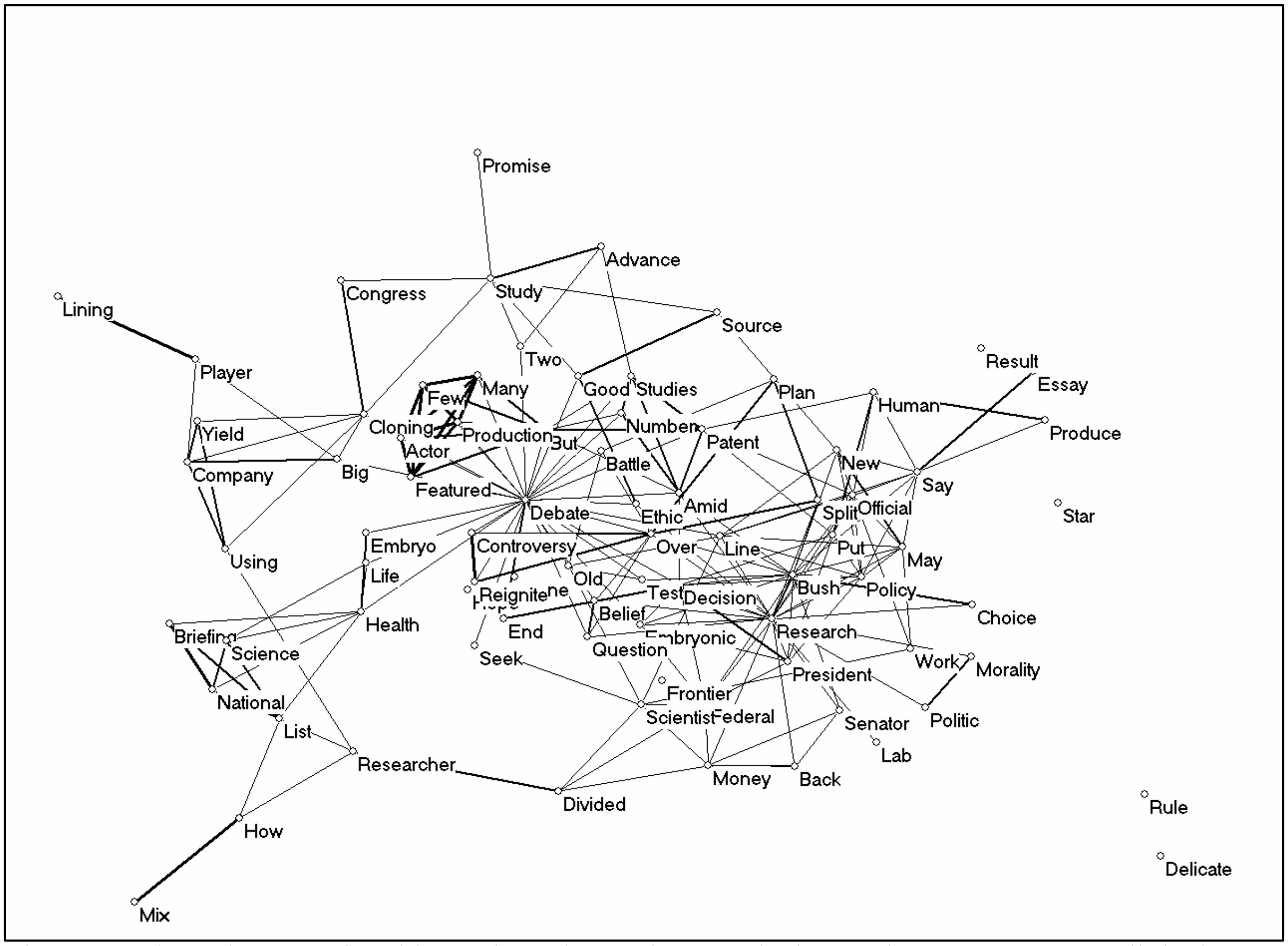

**Figure 10:** The cosine map of 81 title words used more than once in the 127 documents on stem cells in the *New York Times* in 2001 (cosine ≥ 0.1).

Inspection of Figure 10 shows us that the debate in the newspaper focused on the political agenda. The word ‘debate’ has the central position of a star in the network. One main cluster of words shows a representation of Bush’s position, with words such as ‘President,’ ‘Bush,’ ‘official,’ ‘policy,’ and ‘decision,’ and on the other side the various aspects of the topic are reflected in words like ‘embryo,’ ‘life,’ ‘health,’ and ‘science.’ Some words (e.g., ‘potential’) that held a central position in Bush’s speech, are not among the words that play a role in the structure of communication in the newspaper. Instead, the popularization of the issue for wider audiences draws from a wide variety of other relevant topics such as cloning.

In the newspaper, the words “stem cells” function as a metaphor that provides a reference to one of the debates on the national policy agenda. The specificity of word usage in this dataset is lower than in Bush’s argument itself. As in the case of the press releases about *Nature* article on Monarch butterflies, the reflection reduces the codification. In other words, the word usage becomes more metaphorical.

As a third set of texts, we analysed scholarly articles indexed in the *Social Sciences Citation Index* in 2001 with the words “stem cells” in their titles. The semantic map is based on the 41 words that occurred more than once in the titles of 53 documents (Figure 11).

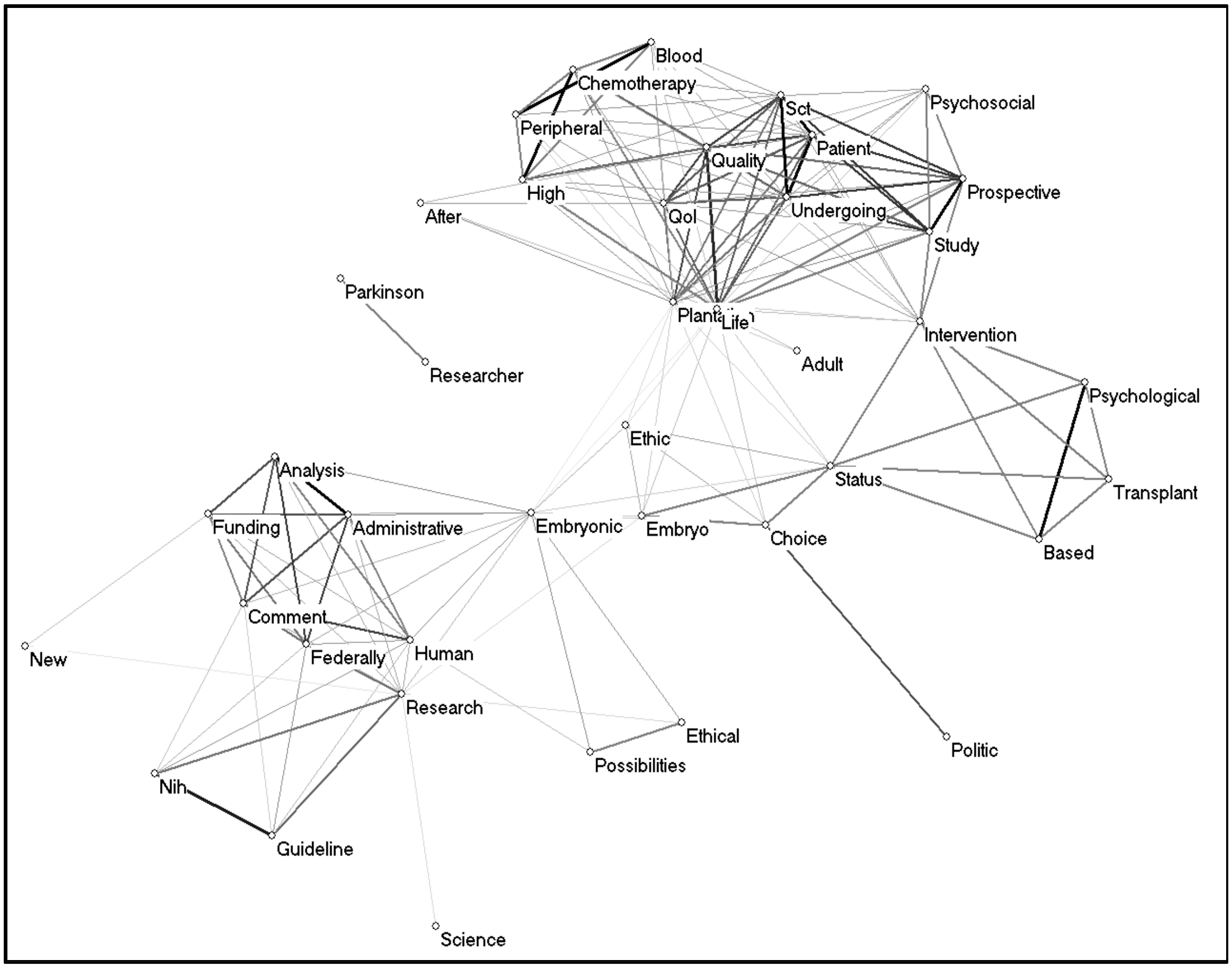

**Figure 11:** The cosine map of 41 title words used more than once in the 53 documents on stem cell in the *Social Science Citation Index* 2001 (cosine ≥ 0.1).

In Figure 11, scholarly articles are differently codified into discourses: medicine, effects on patients, administration science (regulation), and ethics are all represented in the map. Specific words, such as 'status,' 'embryonic,' and 'intervention,' tie some of these clusters together. The different paradigms in these sciences operate as different codifiers. In other words, the words "stem cells" have a specific meaning in these different discourses, which counter-actacts upon the metaphorical function of these words in the public domain. Thus, we observe how the words "stem cell" can function as a metaphor in one context and as a diaphor in another.

In conclusion, the techniques presented here allowed us to automatically map the different degrees of codification of the words "stem cell" across the various places in the continuum between the sciences and society. However, this continuum is highly structured by interfaces. While the words were provided with very specific meanings in the case of Bush's speech as a single document, the documents in the *New York Times* use “stem cells” as a metaphor in the ongoing political debate. The scholarly discourses—as reflected in the titles of contributions to the *SSCI* indexed literature—are highly codified.

**Discussion**

In order to study semantic differences among individual texts and document sets, we have applied automated analysis of *(co-)words in contexts* to three different case studies. First, we analysed semantic differences in the frames of five documents because this allowed us to compare our results with those of an independent and previous content analysis. Second, we followed semantic changes over time in the structural dynamics of the co-word networks of Frankenstein foods. Third, we mapped semantic differences across various domains relevant for the debate on stem cells. In all of these case studies, we were able to map word meanings of the words independently from any *a priori* definition in a scheme or code book by taking into account both the relations of the words and the positions of these words in the distribution of relations. We specified *ex ante* only the three flagship words of the scientific controversies (i.e., “Monarch butterflies,” “Frankenfoods,” and “stem cells”).

Our approach differs from that suggested by Callon *et al*. (1991), Ruiz-Baños *et al*. (1999) and Stegmann & Grohmann (2003) because these authors analysed co-occurrences amongst a set of key words. Their constructivist focus is on comparing the strength of the links, while our focus is on the structure in the *constructed* system of communication; specifically, how the words are positioned as a result of the linking and non-linking among them. Furthermore, we were able to overcome some of the problems of the co-word analysis in the sociology of translation: first, our method is not limited to the key words assigned to the text documents; second, our technique can be applied to large sets without reducing the information content to the symmetrical co-occurrence matrix. The variables of the asymmetrical matrix of documents versus words can be considered as word vectors and accordingly we can use the vector-space model (Salton & McGill, 1983), while the normalization of the co-occurrence matrix has remained debatable (Leydesdorff & Vaughan, forthcoming).[10] Third, by focusing not only on the relations between the words but also on their positions, we are able to measure the meanings of the co-words in their specific contexts.

[10] In earlier work, researchers in the ‘sociology of translation’ (e.g., Callon *et al*., 1983, 1986, 1991)

In the debate on the effects of GM-pollen on Monarch butterflies, we were able automatically to filter out some of the semantic differences constituting the frames of reference distinguished by Nucci (2004) on the basis of a content analysis. The semantic maps showed additional topics used in these domains, i.e., they demonstrated the structures in the contexts of communication. However, the semantic analysis could not inform us about the arguments made in these documents because the arguments belong to the pragmatic dimension of the communication. An analyst may have to focus on certain aspects in the semantic maps before the maps become meaningful. For example, we concentrated on the positioning of the words 'pollen' and 'Monarch' across the documents for the construction of our narrative.

In the case study of "Frankenfoods," we showed the dynamics of the network of co-occurring words over time. These networks changed from an emerging topic in 1996, headed for a clearly delineated and highly structured network in 1999, to a dispersed network of words in 2003. The metaphor of Frankenfoods functioned as an anti-codifier which blurs codified distinctions among domains over time. Further research is needed to specify whether other metaphors function similarly. Finally, in the case study focusing on "stem cells" we were able to show how the scientific and public contexts operate differently. The degree of codification is dependent on the context: a single text document is carefully constructed—therefore dense in its relations—and highly codified; a set of documents can be less codified and less densely packed. In the case of the *Social Sciences Citation Index*, however, the further differentiation according to disciplinary boundaries provided another structure. The word structure is highly organized by the scholarly reflection.

In this study, we used pragmatic cut-off levels of approximately one hundred words for the semantic maps. A threshold was set in the case of 2003 *AltaVista* data because of the huge number of documents retrieved and the limits to visualization on a screen. We are aware that this introduces error as did various other decisions, such as using a standardized stop word list across

---

argued in favour of using the Jaccard Index, while more recently, the so-called Equivalence Index has been used (e.g., Stegmann & Grohmann, 2003). The Equivalence Index is identical to the quotient between observed and expected values which is used in the computation of chi-square values. The measure has also been advocated also for the normalization of co-citation and co-authorship networks domains, etc. It is technically possible to include large numbers of words in the analysis, and the resulting semantic maps can be made so that one is able to zoom in and out on the computer screen. One can also refine the use of stop words and make this selection domain-specific by using, e.g., Bookstein *et al.*'s (1995) statistical approaches.

Our main argument was at another level: we wished to show that the *position* of words in semantic fields can be used as indicators of their meaning. We used relatively straightforward standardizations of techniques in order not to load the article with methodological details. The two concepts of positions and relations are associated because the relations add up and interact in a non-linear way; the positions are generated and stabilized within networks of relations (Burt, 1982). The analysis of relations between positions, however, requires the specification of a reflexive mechanism. In this study, we explored metaphors and diaphors as such mechanisms. In general, the reflexive layer introduces a third system of reference that may reduce or aggravate the uncertainty in the network (Leydesdorff, 2003; Jakulin & Bratko, 2004).

Our results were based on normalizing the number of words included in the analysis without paying attention to the relative weights of the sets in terms of the number of documents or paragraphs within each unit of analysis. There is need for further research into normalizing the numbers of the units of analyses (Theil, 1972). For example, one might consider varying the size of the vertices proportionally to the number of the units of analysis involved (Leydesdorff, 2005).

## Conclusions

In conclusion, techniques for mapping the semantic meanings of co-words in contexts are suitable for automated filtering of the meanings of the words in their different domains of use, over time as well as across varying sets of texts. Focusing on specific functions of words—such as metaphors and diaphors—enabled us to specify the context in which these words gain their meanings. This specification enabled us to make the differences in the frames visible, to follow the development of codes of communication over time, and to analyze different degrees of codification used by various sides at the science/society interfaces. Hence, the method can be

(Zitt, 2000; cf. Michelet, 1988; Leydesdorff & Vaughan, forthcoming).

applied to a wide variety of longitudinal studies of science communication as well as comparative studies across the various domains of communication among the sciences and at science/technology/society interfaces (Leydesdorff, 2004). The differences among the domains of use inform us about the variation in the discourses, and about the selections in their respective operations. The methodology can also be used as an alternative to content analysis in the case of large (e.g., electronic) datasets that can no longer be coded manually.

The study contributes to several research traditions that aim to automate the mapping of the dynamics of communications. On the one hand, we were able to operationalize the mapping of the dynamics of knowledge (Maasen and Weingart, 2000). On the other hand, the specification of the context in which the co-words occur takes part within the debates on the sociology of translation (Callon *et al.*, 1983: Callon *et al.*, 1991) and automated content analysis (Krippendorff, 1980/2002: Klein, 2004). The contexts can only be specified if reflexive mechanisms are defined.

In our case studies, two reflexive mechanisms were identified for the function of translation: metaphors and diaphors. Metaphors operate as foci of reflection, while diaphors contribute to the discourse as a distribution of words. The distributions are spatially arranged in networks. These networks are interfaced at each moment of time, but they contain codes which develop over time. Thus, there is both a dynamic and a synchronic aspect to the contexts. The operation of structures at each moment in time and their stabilization over time can be expected to lead to the globalisation or the decay of the knowledge base of codifications, due to the meta-stabilities that can be expected in the interactions among the differently codified subdynamics of the communication (MacKenzie, 2001; Hellsten *et al.*, forthcoming).

**Annex 1: The List of Documents Used in the Case Study On Monarch Butterflies (see also Nucci 2004)**

Losey, John E., Rayor, Linda S., Carter, Maureen E. 'Transgenic Pollen harms monarch larvae.' *Nature* 399:214, 20 May 1999.

The Cornell University press release, entitled 'Engineered corn can kill monarch butterflies', released on May 19, 1999.

The UCS document, entitled 'Toxic pollen threatens monarchs – gene-altered corn may harm beloved butterfly' published in the May 1999 issue.

The Greenpeace document entitled 'Monsanto and Novartis genetically engineered maize harms butterflies – Greenpeace calls for a ban' published on May 20, 1999.

The BIO document, entitled 'BIO responds to Nature report on Bt threat on Monarch butterflies', published on 12 August, 1999.